\documentclass[letterpaper, 10 pt, conference]{ieeeconf}
\IEEEoverridecommandlockouts 
\usepackage[table]{xcolor}
\usepackage{hyperref}
\usepackage{graphicx}
\usepackage{amsmath,amssymb}
\usepackage{cite}
\usepackage{float}
\usepackage{array}
\usepackage[caption=false,font=footnotesize]{subfig}
\usepackage{todonotes}
\usepackage{caption}

\begin{document}

\renewcommand{\vec}{\boldsymbol}

\graphicspath{{img/},{fig/}}

\title{\LARGE \bf
Using the Buckingham $\pi$ Theorem for Multi-System Transfer Learning:  a Case-study with 3 Vehicles Sharing a Database




}

\author{ William Therrien, Olivier Lecompte and Alexandre Girard$^{1}$
\thanks{$^{1}$All authors are with the Department of Mechanical Engineering, Universite de Sherbrooke, Qc, Canada {\tt\small  Contact: alex.girard@usherbrooke.ca }}
}%

\maketitle
\thispagestyle{empty}
\pagestyle{empty}

\begin{abstract}
Many advanced driver assistance schemes or autonomous vehicles controllers are based on a motion model of the vehicle behaviour, i.e. a function predicting how will the vehicle react to a given control input. Data-driven models, based on experimental or simulated data, are very useful especially for vehicles difficult to model analytically, for instance, ground vehicles for which the ground-tire interaction is hard to model from first principles. However, learning schemes are limited by the difficulty of collecting large amounts of experimental data or having to rely on high-fidelity simulations. This paper explores the potential of an approach that uses dimensionless numbers based on Buckingham's $\pi$ theorem to improve the efficiency of data for learning models, with the goal of facilitating knowledge sharing between similar systems. A case study using car-like vehicles compares traditional and dimensionless models on simulated and experimental data to validate the benefits of the new dimensionless learning approach. Preliminary results with the case study presented show that this new dimensionless approach could accelerate the learning rate and improve the accuracy of the model prediction when transferring the learned model between various similar vehicles. Prediction accuracy improvements with the dimensionless scheme when using a shared database, that is, predicting the motion of a vehicle based on data from various different vehicles, was found to be 480 \% more accurate for predicting a simple no-slip maneuver based on simulated data and 11 \% more accurate to predict a highly dynamic braking maneuver based on experimental data. A modified physics-informed learning scheme with hand-crafted dimensionless features was also shown to increase the improvement to precision gains of 917 \% and 28 \% respectively. A comparative study also show that using the Buckingham's $\pi$ theorem is a much more effective preprocessing step for this task than principal compoment analysis (PCA) or simply normalizing the data. These results show that the use of dimensionless variables is a promising tool to help in the task of learning a more generalizable motion model for vehicles, and hence potentially taking advantage of the data generated by fleets of vehicles on the road even tough they are not identical.
\end{abstract}

\section{Introduction}

The typical approach to solving motion control problems for autonomous vehicles and other types of robotic systems is to use a physics-based kinematic or dynamic model of the system to plan trajectories and establish feedback laws \cite{amer_modelling_2017}\cite{paden_survey_2016}\cite{katrakazas_real-time_2015}. With the rise of efficient machine learning algorithms, much effort has been spent in using learning in planners and controllers so that autonomous vehicles can learn from experience and improve over time\cite{aradi_survey_2022}\cite{crites_improving_nodate}\cite{di_survey_2021}. However, major difficulties limit the success of the learning scheme when it comes to controlling real physical platforms. One of the major challenges is the difficulty of collecting the large amount of experimental data required for advanced learning schemes such as deep-learning \cite{di_survey_2021}\cite{lake_building_2017}. A popular solution is to generate data using high-fidelity simulations instead of experiments. Nevertheless, this approach also has multiple challenges, such as the difficulty of capturing certain behaviours in a simulator and transferring the results to the real world \cite{liu_digital_2022}. All in all, data efficiency is a critical bottleneck.

\begin{figure*}[ht]
    \centering
    \subfloat[Small vehicle]{
        \includegraphics[width=0.25\textwidth]{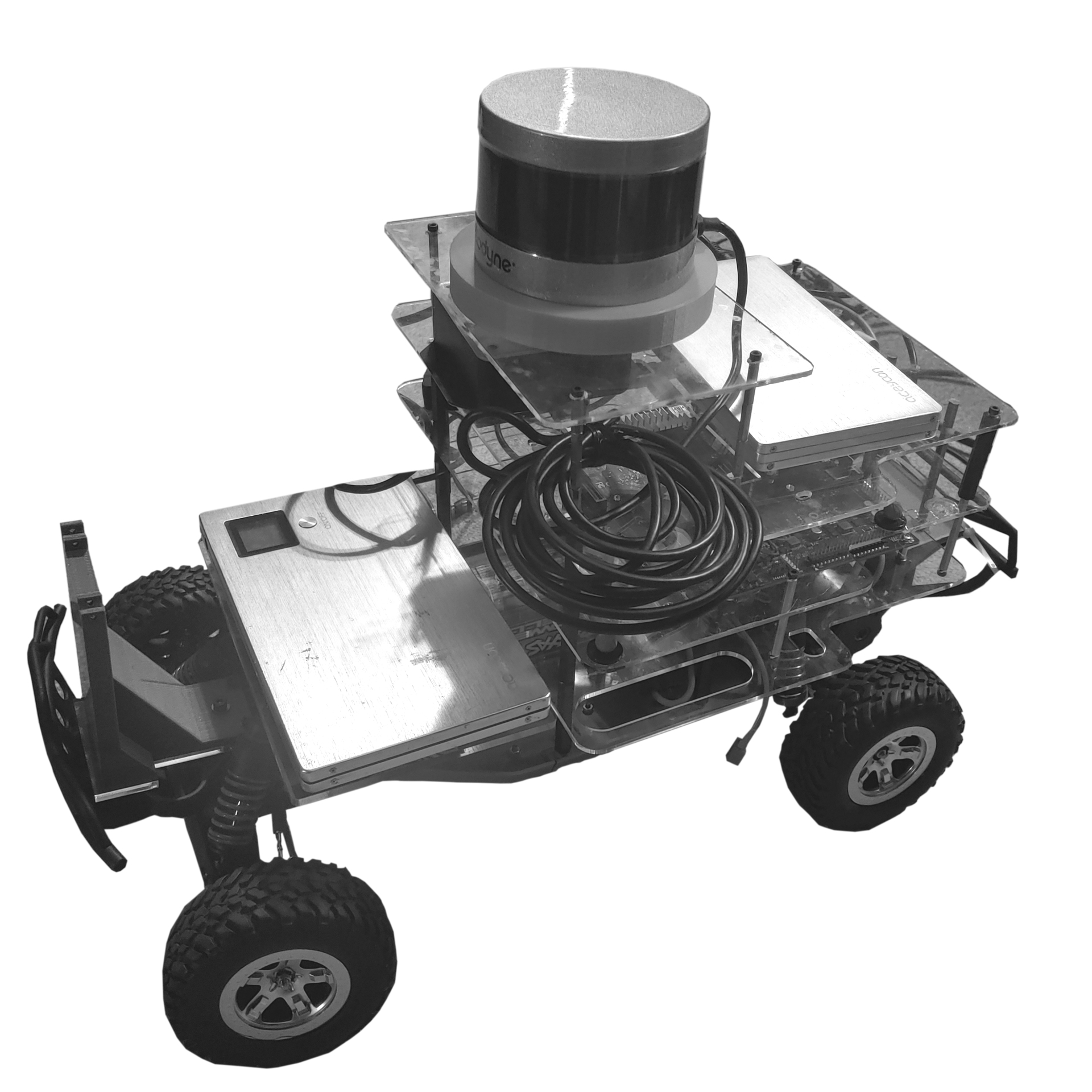} 
        \label{fig:racecar_pic}
        }
    \subfloat[Long vehicle]{
        \includegraphics[width=0.35\textwidth]{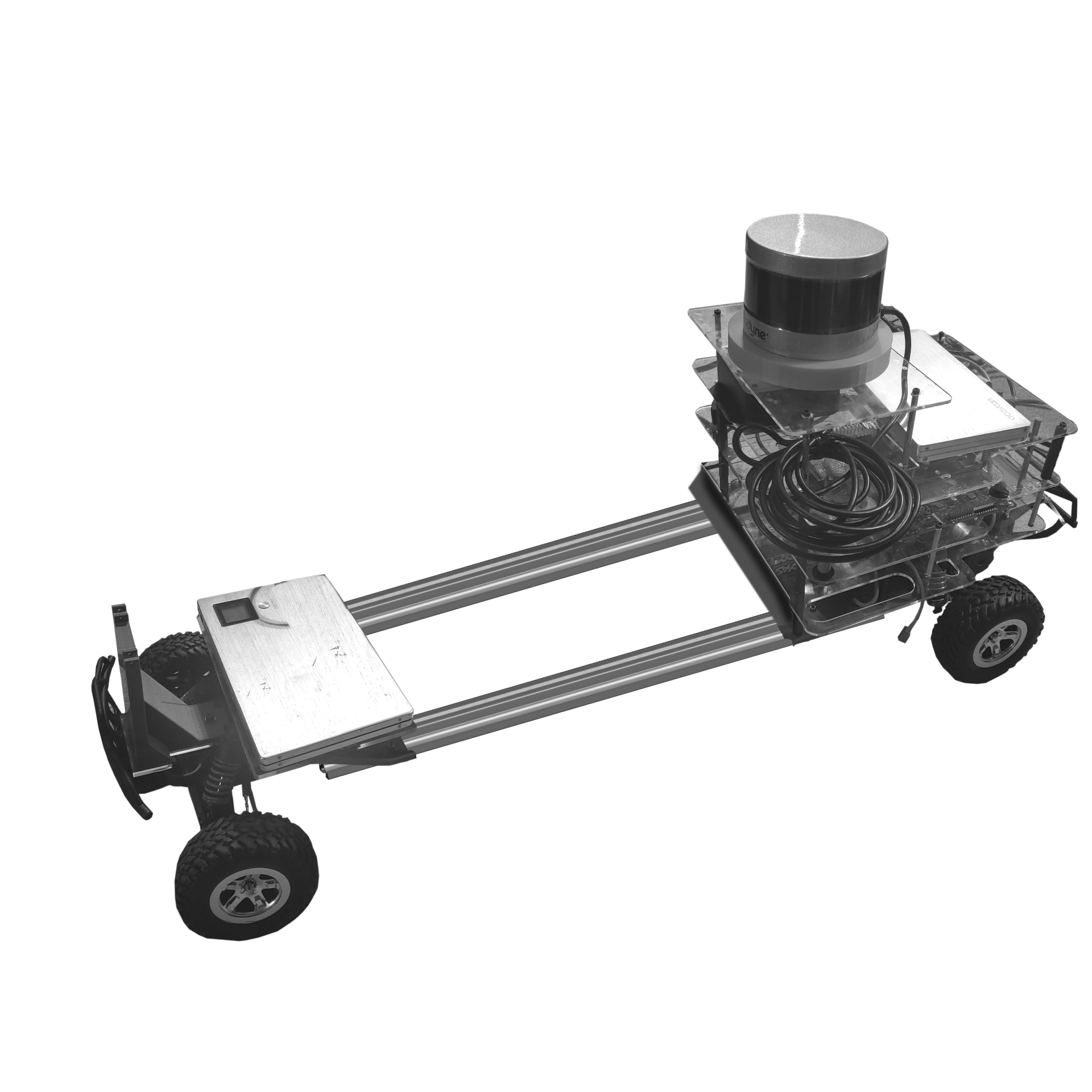} 
        \label{fig:limo_pic}
        }
    \subfloat[Large vehicle]{
        \includegraphics[width=0.35\textwidth]{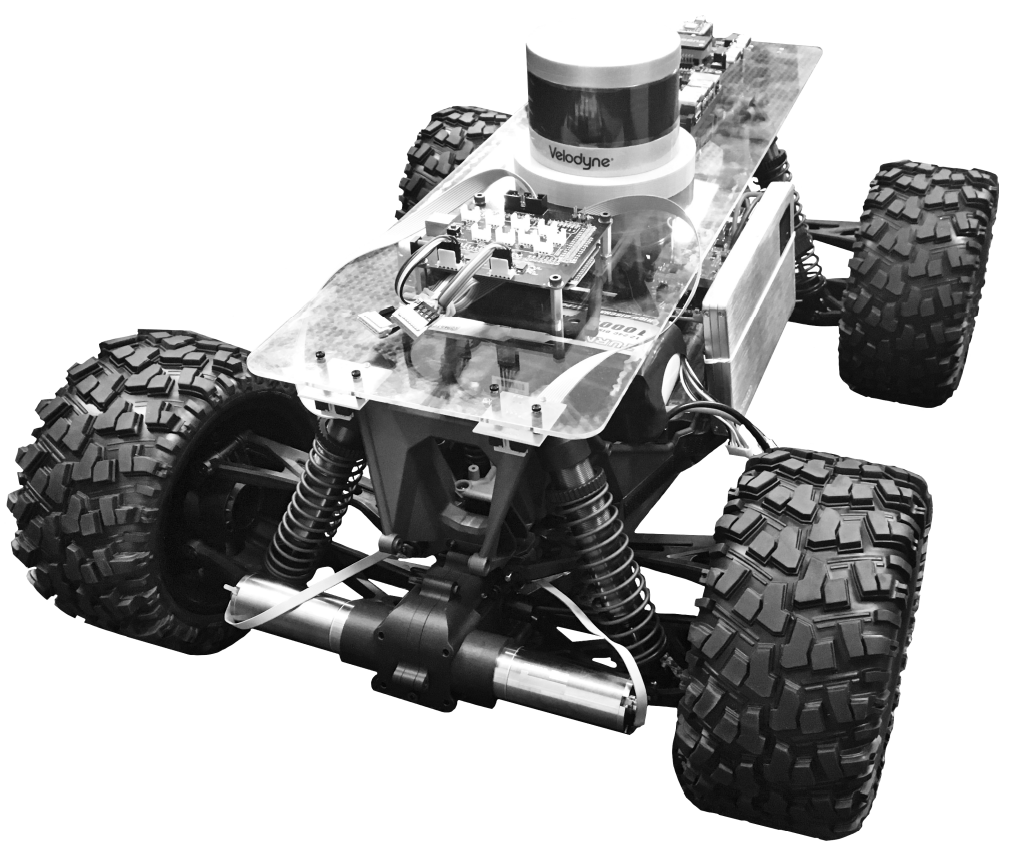} 
        \label{fig:Xmaxx_pic}
        }
    \caption{Small scale platforms of various dimensions used for experimental validation. }
    \label{fig:vehicles}
\end{figure*}

Sharing learned data and policies across multiple systems could help solve the problem of collecting huge amounts of data. For example, instead of having to start learning from scratch, a centralized database for a fleet of vehicles would enable the newly deployed vehicle to use knowledge based on thousands of hours of motion data collected by hundreds of vehicles. However, this is difficult to achieve when the studied systems do not share identical characteristics and, therefore, the appropriate control policy differs. This paper investigates the potential of leveraging dimensionless number and the Buckingham $\pi$ theorem to help learning more general models. 
Section \ref{sec:rel} presents some related work and discusses the original contribution of this paper. Section \ref{sec:casestudy} defines the problem used as the case study in this paper. Section \ref{sec:simulation} presents the learning results based on simulated data and section \ref{sec:exp} presents the learning results based on an experimental validation with three small scale vehicles shown at Fig. \ref{fig:vehicles} .

\section{Background}
\label{sec:rel}
Multi-task transfer learning has recently received significant research attention \cite{dey_learning_2021}\cite{andrychowicz_learning_2020}\cite{nagabandi_learning_2019}.
This paper focuses on multi-system transfer where the objective is to share knowledge between similar systems that accomplish the same task.  
Specifically for multi-system transfer, an interesting approach is presented in \cite{dasari_robonet_2020} where a shared database can outperform a robot-specific training approach that uses 4x-20x more data. In \cite{sorocky_experience_2020}, the authors address the problem of inappropriate transfer, which can result in a decrease in performance. By assessing how similar two systems are, it is possible to know with which systems the transfer learning would be most beneficial. The closest related work is probably \cite{chen_hardware_2018}, where the authors tackle the problem directly by guiding the policy network with a representation vector built from the information on the hardware. The goal is similar in this work, but the approach is different: instead of two vectors, one containing the states and the other containing the characteristics of the hardware, dimensional analysis is used to build a single dimensionless vector using the Buckingham $\pi$ theorem. 

Dimensional analysis \cite{bertrand_sur_1878} \cite{rayleigh_viii_1892}\cite{buckingham_physically_1914}, i.e. the Buckingham $\pi$ theorem, is a useful tool to simplify input-output relationships that involve physical quantities. It is used extensively in the field of fluid mechanics to minimize the number of experiments required to characterize a behaviour. There is a recent renewed interest in using dimensional analysis in the context of learning. It was proposed to pre-process data with the Buckingham $\pi$ theorem to improve robustness and accuracy of learned complex turbulent flow dynamics \cite{fukami_robust_2021}\cite{Fukami_2024}. Also, automated methods were proposed to identify the most relevant dimensionless groups of dynamical systems \cite{bakarji_dimensionally_2022}\cite{xie_data-driven_2022}. In other works, Buckingham $\pi$ neural network were proposed, and shown to generalize better than dimensional regular neural network for predicting simple simulated physical systems, a thermal management systems in \cite{OPPENHEIMER2023111810} and a vegetation model in \cite{villar2022dimensionless}. Furthermore, in \cite{OPPENHEIMER2023111810} results shown that for the conducted experiments, using the Buckingham $\pi$ method was more efficient than other reduction technique like principal component analysis and autoencoders. Finally, clustering dimensionless learning was proposed as an approach to identify regime of behaviours \cite{ZHANG2024116728}. All in all, many very interesting works have shown the potential of using the Buckingham $\pi$ theorem for improving generalization of machine learning models. However, since most of the results are for very simple simulated experiments and in the context of fluid mechanics prediction, many questions still remain unanswered regarding how such techniques would perform in other fields and with experimental data. In this paper, we present a case study to specifically explore the potential of such approaches in order to provide a potential solution to the problem of transfer learning for motion model of car-like vehicles.

Dimensional analysis it still very seldomly explored in the field of motion control and robotics, to our knowledge only a few projects explored that concept. In the case of an in-lane collision avoidance system, \cite{singh_nondimensionalized_2016} created dimensionless indices, based on a vehicle's relative longitudinal velocity, road surface conditions and terminal lateral distance, which were used in the selection of the most efficient maneuver and timing for an intervention. The use of these indices was shown to be computationally efficient. The authors of \cite{luo_kinematic_2021} proposed an improved method for kinematic calibration based on dimensionless error mapping matrices (EMMs). Simulation results show that the residual pose errors with the proposed dimensionless EMMs were lower than with the conventional EMM in various units. Finally, \cite{math12050709} presents the concept of dimensionless policies. It discusses the notion that similar dynamic systems must have an equivalent optimal policy when expressed in dimensionless form and shows how to exactly transfer a motion control policy. However, it does not provide results regarding the quality of transfers when the similarity condition is not exactly met, which is studied in this paper with a case-study.

The contribution of this paper is to demonstrate the advantage of learning schemes leveraging the Buckingham $\pi$ theorem, i.e. dimensional analysis, specifically for transferring motion models between car-like vehicles. While the idea of using dimensionless input-output variables to improve generalization of data-driven model is not new, the novelty of this paper is to evaluate its performance for the specific application of sharing data and model between car-like vehicles. Furthermore, it also provides an important contribution by showing that the advantage of the implemented technique was also present in a real-life experiment involving small autonomous car-like robots, as opposed to previous works where most results were based on simulated data in ideal conditions. %

\section{Case-study: predicting the braking behaviour of cars}
\label{sec:casestudy}

The main goal of this paper is to validate whether a learning scheme can benefit from a dimensionless mapping of its inputs and outputs. These benefits could take the form of an increase in learning speed, a reduction in mean absolute error (MAE) and the possibility of sharing knowledge between similar systems. To do so, three different learning schemes are compared using their respective MAEs; all are based on optimized distributed gradient boosting (XGBoost) models trained with simulated and experimental data. The first one, the baseline of the comparison, is using directly the XGBoost model on the physical input and output data without any prepossessing. The second one is based on transforming the input and output data into dimensionless variables, based on the Buckingham $\pi$. The third approach is also based on the Buckingham $\pi$ theorem but with additional dimensionless inputs (hand-crafted features based on expert knowledge of the system's physics). Inspired by the challenges of learning good control policies for emergency braking maneuvers under various ground conditions (snow, ice, etc.), such as explored in \cite{lecompte_experimental_2022}, the case study used in this paper is the prediction of the final relative position of a vehicle after a sudden emergency braking maneuver, based on initial conditions, control inputs (braking and steering), and environmental parameters. The learned model could then be used in a planning and control pipeline to select the best maneuver in an emergency braking situation for instance. To simplify the case study a few assumptions are made. Firstly, the control inputs are constant throughout a maneuver, i.e. the steering angle $\delta$ and braking action (constant deceleration rate $a$ imposed on both rear wheels) are decided once at the start of the maneuver and remain constant until the vehicle comes to a complete stop.  Also, only the final position of the vehicle is predicted, not the complete trajectory. Furthermore, the environment is always a horizontal surface and only 2D planar motion is taken into account. The simulation is based on a kinematic bicycle motion model, while for the experimental data the dynamics of the system is taken into account, neglecting any roll or pitch effects and considering the complex soil-tire relationship as a constant coefficient of friction $\mu$. Table \ref{expVari} shows the main variables involved in this case study.

Three small scale car-like vehicles of various dimension are used for both validation processes (simulation and experimentation). Vehicle 1, shown in figure \ref{fig:racecar_pic}, is a modified 1:10 scale platform based on a Traxxas SLASH. The vehicle 2, shown in figure \ref{fig:limo_pic}, is also a modified 1:10 scale platform based on a Traxxas SLASH with the chassis modified to have a longer wheelbase. Finally, vehicle 3, shown in figure \ref{fig:Xmaxx_pic}, is a modified 1:5 scale platform based on a Traxxas X-MAXX. They will be referred to as \textit{small vehicle}, \textit{long vehicle} and \textit{large vehicle} respectively. Their specifications are summarized in table \ref{specs}. 


\begin{table*}[htbp]
   \centering 
   \caption{Variables used throughout this paper} 
   \label{expVari}
   \begin{tabular}{c p{8.0cm} c }
   \hline
   \hline \noalign{\smallskip} \noalign{\smallskip} \noalign{\smallskip}
   \textbf{Variables} & \textbf{Descriptions} & \textbf{Units [Dimensions]} \\ \noalign{\smallskip} \noalign{\smallskip} \hline \hline \noalign{\smallskip} 
   \multicolumn{3}{c}{\textbf{State variables}}\\ \noalign{\smallskip}  \hline \hline \noalign{\smallskip} 
   $X$ & X-axis position of the vehicle in the world frame & m [L]\\ \noalign{\smallskip} \hline \noalign{\smallskip}
   $Y$ & Y-axis position of the vehicle in the world frame & m [L]\\  \noalign{\smallskip} \hline \noalign{\smallskip}
   $\theta$ & Yaw of the vehicle in the world frame & rad  \\ \noalign{\smallskip} \hline\hline \noalign{\smallskip}
   \multicolumn{3}{c}{\textbf{Environment related variables}}\\ \noalign{\smallskip}  \hline\hline  \noalign{\smallskip} 
   $\mu$ & Friction coefficient wheels/road & -  \\ \noalign{\smallskip} \hline \noalign{\smallskip}
   $v$ & Longitudinal velocity of the vehicle & m/s [LT$^{-1}$] \\ \noalign{\smallskip} \hline \noalign{\smallskip}
   $g$ & Gravitational acceleration & m/s$^2$ [LT$^{-2}$]  \\ \noalign{\smallskip} \hline \hline \noalign{\smallskip}
   \multicolumn{3}{c}{\textbf{maneuvers related variables}}\\ \noalign{\smallskip}  \hline \hline \noalign{\smallskip} 
   $a$ & Deceleration of the wheel & m/s$^2$ [LT$^{-2}$] \\ \noalign{\smallskip} \hline \noalign{\smallskip}
   $\delta$ & Steering angle of front wheels & rad  \\ \noalign{\smallskip} \hline \hline \noalign{\smallskip}
   \multicolumn{3}{c}{\textbf{Vehicles related variables}}\\ \noalign{\smallskip}  \hline \hline \noalign{\smallskip} 
   $N_f$ & Normal force on front wheels & N [MLT$^{-2}$] \\ \noalign{\smallskip} \hline \noalign{\smallskip}
   $N_r$ & Normal force on rear wheels & N [MLT$^{-2}$] \\ \noalign{\smallskip} \hline \noalign{\smallskip}
   
   $l$ & Length between vehicle's axles & m [L]  \\  \noalign{\smallskip} \hline \hline \noalign{\smallskip}
   \end{tabular}
\end{table*}

\begin{table*}[htbp]
   \centering 
   \caption{Specifications of the three experimental vehicles} 
   \label{specs}
   \begin{tabular}{l c c c }
   \hline 
   \hline \noalign{\smallskip} \noalign{\smallskip} \noalign{\smallskip}
   & Wheel base length & Weight on front wheels (static) & Weight on back wheels (static)\\ \noalign{\smallskip} \hline 
   \textbf{Vehicle} & $l$ [m]  & \textbf{$N_f$} [N] & \textbf{$N_r$} [N]\\ \noalign{\smallskip} \noalign{\smallskip} \hline \hline \noalign{\smallskip} 
   \textbf{Small} & 0.345 & 37.77 & 28.84\\ \noalign{\smallskip} \hline \noalign{\smallskip}
   \textbf{Long} & 0.853 & 22.74 & 52.89\\  \noalign{\smallskip} \hline \noalign{\smallskip}
   \textbf{Large} & 0.475 & 71.12 & 71.12 \\ 
   \noalign{\smallskip} \hline \hline \noalign{\smallskip}
   \end{tabular}
\end{table*}


\section{Learning with simulated data}
\label{sec:simulation}

\subsection{Simulator presentation}
This section presents an analysis in which the proposed learning approach is evaluated using data generated in a simplified simulated environment. The simulation was based on the kinematic bicycle model frequently used in the literature. The non-linear equations for the bicycle model are shown at \ref{kin}. 

\begin{equation}\label{kin}
\begin{bmatrix}
\Dot{X}\\
\Dot{Y}\\
\Dot{\theta}\\
\Dot{v}\\
\end{bmatrix}
=
\begin{bmatrix}
vcos(\theta)\\
vsin(\theta)\\
\frac{vtan(\delta)}{l}\\
a\\
\end{bmatrix}
\end{equation}

For the three vehicles presented in Figure \ref{fig:vehicles}, 500 simulations were performed with the initial values and control input shown in Table \ref{values}, that is, all combinations of 50 initial speed levels, 10 levels of braking, and 11 steering positions. All simulations start with the vehicle at position [\textit{X},\textit{Y}] = (0,0), orientation $\theta$ = 0 rad, and initial speed $v_i$. The maneuver starts immediately with deceleration $a$ and steering angle $\delta$. The simulation stops when the vehicle reaches zero velocity and the final position and orientation are noted.

\begin{table}[ht]
   \centering 
   \caption{Values for the simulations for all 3 vehicles} 
   \label{values}
   \begin{tabular}{c c c}
   \hline 
   \hline \noalign{\smallskip} \noalign{\smallskip} \noalign{\smallskip}
     $v_i$ & $a$ & $\delta$ \\ (m/s)& (m/s$^2$)& (rad)\\ \noalign{\smallskip} \noalign{\smallskip} \hline \hline \noalign{\smallskip} 
    From 0.1 & From -0.1 g & From 0.0000\\ 
    to 5.0 & to -1.0 g & to 0.7854 \\
    by 0.1 & by -0.1 g & by 0.0785 \\\noalign{\smallskip} 
   \hline\noalign{\smallskip} 
    50 values & 10 values & 11 values \\\noalign{\smallskip} 
   \noalign{\smallskip} \hline \hline \noalign{\smallskip}
   \end{tabular}
\end{table}

\subsection{Learning model types for the simulation}

Here the detail of a baseline reference prediction scheme and two novel dimensionless prediction scheme are presented.

\subsubsection{Traditional dimensionalized learning model} 
The process of integrating the nonlinear equations describing the kinematic bicycle model used for the simulation discussed in section \ref{kin} until the vehicle comes to a stop, can be written in the form of input-relationship as given by \eqref{fonc_sim}.

\begin{equation}\label{fonc_sim}
\begin{bmatrix}
X \\ Y \\ \theta
\end{bmatrix}
=
f \Bigl(v_i,a, \delta, l \Bigr)
\end{equation}

\begin{figure}[ht]
\begin{center}
\includegraphics[width=0.99\linewidth]{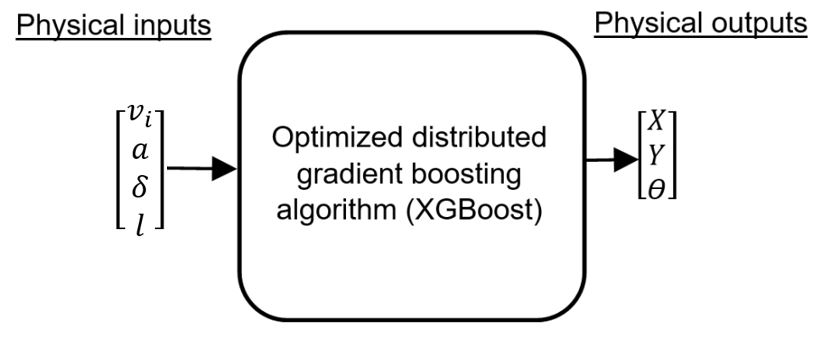}
\caption{Baseline learning scheme for simulated data}\label{fig:PM_kine}
\end{center}
\vspace{-10pt}
\end{figure}

As a baseline learning scheme for this evaluation, the XGBoost algorithm is used directly to learn a black-box input-output model based on the data. Thus, the dimensionalized parameters for the initial velocity $v_i$, the decelaration \textit{a}, the steering angle $\delta$ and the length of the wheelbase \textit{l} are used to predict the final pose [\textit{X}, \textit{Y}, $\theta$] of the vehicle. This traditional dimensionalized learning scheme is illustrated in figure \ref{fig:PM_kine}

\subsubsection{Buckingham $\pi$ theorem based model}

According to Buckingham $\pi$ theorem \cite{buckingham_physically_1914}, a relationship with \textit{N} dimensional variables and \textit{P} independent dimensions, can be restated as a relationship between \textit{m} = \textit{N} - \textit{P} dimensionless numbers. Here, each line of equation \eqref{fonc_sim}, represent a relationship between \textit{N} = 5 variables and \textit{P} = 2 fundamental dimensions (length [L] and time [T]) involved. Each line of equation \eqref{fonc_sim} can thus be restated as a relationship between 3 dimensionless variables, leading the following:
\begin{align}\label{fonc_sim_dim_pi}
\begin{bmatrix}
\pi_1 \\ \pi_2 \\ \pi_3
\end{bmatrix}
&=
f \Bigl( \pi_4 , \pi_5 \Bigr)
\end{align}
Here, the dimensionless variables are selected using the distance between the two axles $l$ [M] and the initial velocity of the vehicle $v_i$ [LT$^{-1}$] as repeated variables in the dimensional analysis:
\begin{align}
\pi_1 = \frac{X}{l} \quad 
\pi_2 = \frac{Y}{l} \quad 
\pi_3 = \theta \quad 
\pi_4 = \frac{al}{v_i^2} \quad 
\pi_5 = \delta
\end{align}
The new dimensionless relationship to predict is thus now, restating \eqref{fonc_sim} with the selected dimensionles varibles, the following input-output relationship:
\begin{align}\label{fonc_sim_dim}
\begin{bmatrix}
\frac{X}{l} \\ \frac{Y}{l} \\ \theta
\end{bmatrix}
&=
f \left(\frac{al}{v_i^2}, \delta \right)
\end{align}

\begin{figure*}[ht]
\begin{center}
\includegraphics[width=0.7\linewidth]{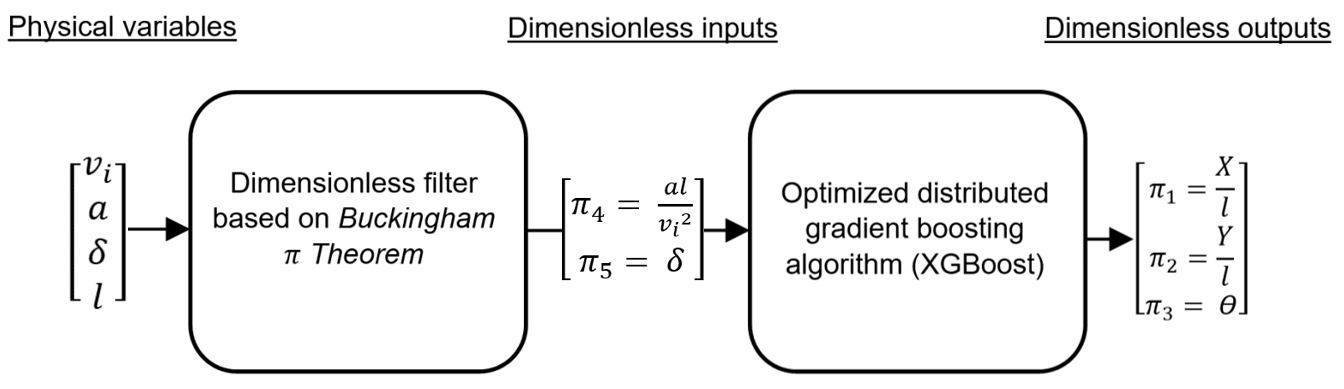}
\caption{Buckingham $\pi$ theorem based model for simulated data}\label{fig:BM_kine}
\end{center}
\vspace{-10pt}
\end{figure*}

In is interesting to note that, in dimensionless form the motion behaviour model has an reduced input space of 2 dimension instead of 4 initially with equation \eqref{fonc_sim}. The dimensionless learning scheme consist of using the same XGBoost algorithm, but on input-output data converted into those dimensionless variables. As illustrated in Figure \ref{fig:BM_kine}, the scheme can be seeing as a prepossessing of the data using intelligent scaling based on dimensional analysis.

\subsubsection{Augmented Buckingham $\pi$ theorem based model}

A secondary question that is explored in this manuscript is the impact of adding supplementary physically relevant hand-crafted dimensionless variables as inputs to a learning algorithm. This scheme will be referred to as the \textit{ Augmented Buckingham $\pi$ theorem-based model}. Here, an arbitrary dimensionless number $\pi_6$ that is constructed based on the initial angular velocity is included:
\begin{equation}\label{eq:pi6}
\Dot{\theta} = \frac{v\tan(\delta)}{l} 
\quad \quad \Rightarrow \quad \quad
\pi_{6} = \frac{{v^2}\tan(\delta)}{al} 
\end{equation}
The learned input-output relationship is thus modified into the form:
\begin{align}\label{fonc_sim_dim2}
\begin{bmatrix}
\frac{X}{l} \\ 
\frac{Y}{l} \\ 
\theta
\end{bmatrix}
&=
f \Bigl( \pi_4, \pi_5, \pi_6  \Bigr)
=
f \left( \frac{al}{v_i^2}, \; \delta, \; \frac{v^2\tan(\delta)}{al}  \right)
\end{align}

\begin{figure*}[ht]
\begin{center}
\includegraphics[width=0.8\linewidth]{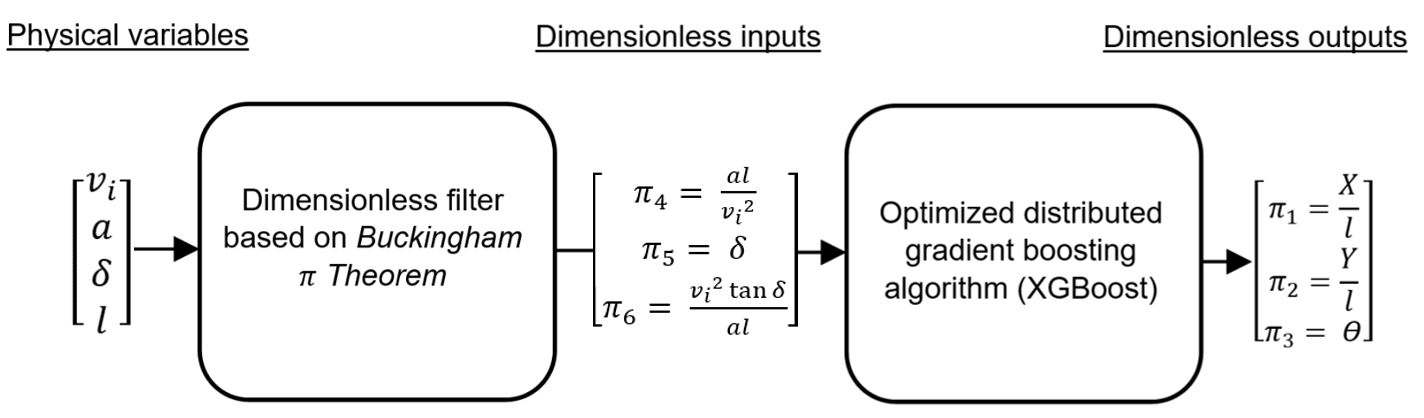}
\caption{Augmented Buckingham $\pi$ theorem based model for simulated data}\label{fig:AM_kine}
\end{center}
\vspace{-10pt}
\end{figure*}

The hypothesis is that adding this hand crafted feature, that should be more closely related to predicting outputs based on first principles, might simplify the learning task and improve the performance of the learning algorithm. The inputs and outputs of the \textit{Augmented Buckingham $\pi$ theorem based model} used to predict the outcome of a maneuver are shown in Figure \ref{fig:AM_kine}.

\subsection{Results}
The results for each model type are examined from two perspectives: 1) the accuracy (MAEs) of self-predictions, cross-predictions and shared-predictions; and 2) the effect of the size of the training set on the convergence of the MAEs, i.e., the learning rates for each situations.

\subsubsection{Mean absolute error} The MAE is one of the most widely used metrics for regression algorithms, since it has the same units as the model's dependent variables, which makes the results easier to understand. Equation \ref{eq:MAE} shows how MAE is calculated, where \textit{n} is the number of tests used, $y_i$ are the actual result values and $\hat{y}_i$ are the predicted values. Here the output values for both dimensionless models are $\pi_1$ = $\frac{X}{l}$, $\pi_2$ = $\frac{Y}{l}$ and $\pi_3$ = $\theta$, $\pi_1$ and $\pi_2$ are multiplied by the wheelbase $l$ of the tested vehicle before computing the MAE so that all results share the same units.

\begin{equation}\label{eq:MAE}
MAE = \frac{1}{n} \sum_{i=1}^n|(y_i-\hat{y}_i)|
\end{equation}

To evaluate the performance of the models, 80$\%$ of the simulated data was used to train a model of each type for the three vehicles, and the remaining 20$\%$ was used to test the models. Table \ref{results_P} contains the results of the traditional model (baseline), the results of the model based on Buckingham $\pi$ theorem are presented in table \ref{results_B} and table \ref{results_A} shows the results of the Augmented Buckingham model. For all three tables, the light gray cells correspond to the results of what we call self-predictions, i.e. predictions that use a vehicle's test data in its own trained model. The white cells correspond to the results of a vehicle's test set in another vehicle's trained model, that will be refer to as cross-predictions. Finally, the dark gray cells are the results of shared predictions. For shared predictions, the test set of a particular vehicle is ran for a model that has been trained with the training data from all three vehicles (i.e. a shared database). 
\begin{table}[ht]
\centering
\caption{MAEs for the traditional baseline model (X and Y are in meters and $\theta$ in rad)} 
\label{results_P}
\begin{tabular}{|c|c|c|c|}
  \hline 
  \textbf{Models} & \textbf{Data Vehicle 1} & \textbf{Data Vehicle 2} &\textbf{Data Vehicle 3}\\ 
  & (Small) & (Long) & (Large) \\\hline \hline

  \textbf{Model} & \cellcolor{gray!25}X: 0.0494  & X: 0.3703 & X: 0.1601 \\
  \textbf{Vehicle 1} & \cellcolor{gray!25}Y: 0.0466 & Y: 0.3071 & Y: 0.1459\\
  (Small) & \cellcolor{gray!25}$\theta$: 0.0587  & $\theta$: 0.9851  & $\theta$: 0.4502\\
  \hline
  \textbf{Model} & X: 0.3653 & \cellcolor{gray!25}X: 0.0316 & X: 0.2622\\
  \textbf{Vehicle 2} & Y: 0.3032 & \cellcolor{gray!25}Y: 0.0257 & Y: 0.2061\\
  (Long) & $\theta$: 0.9919  & \cellcolor{gray!25}$\theta$: 0.0238 & $\theta$: 0.5369 \\
  \hline
  \textbf{Model} & X: 0.1667 & X: 0.2679 & \cellcolor{gray!25}X: 0.0414  \\
  \textbf{Vehicle 3} & Y: 0.1440 & Y: 0.2125 & \cellcolor{gray!25}Y: 0.0362 \\
  (Large) & $\theta$: 0.4599  & $\theta$: 0.5324  & \cellcolor{gray!25}$\theta$: 0.0422\\
  \hline
  \textbf{Model} & \cellcolor{gray!75}X: 0.0467 & \cellcolor{gray!75}X: 0.0433 & \cellcolor{gray!75}X: 0.0514\\
  \textbf{MERGED} & \cellcolor{gray!75}Y: 0.0465 & \cellcolor{gray!75}Y: 0.0368 & \cellcolor{gray!75}Y: 0.0498\\
  (All 3) & \cellcolor{gray!75}$\theta$: 0.0457  & \cellcolor{gray!75}$\theta$: 0.0302 & \cellcolor{gray!75}$\theta$: 0.0431 \\
  \hline
\end{tabular}
\end{table}

\begin{table}[htp]
\centering
\caption{MAEs for the Buckingham $\pi$ theorem based model (X and Y are in meters and $\theta$ in rad)} 
\label{results_B}
\begin{tabular}{|c|c|c|c|}
  \hline 
  \textbf{Models} & \textbf{Data Vehicle 1} & \textbf{Data Vehicle 2} &\textbf{Data Vehicle 3}\\ 
  & (Small) & (Long) & (Large) \\\hline \hline

  \textbf{Model} & \cellcolor{gray!25}X: 0.0204  & X: 0.0144 & X: 0.0125 \\
  \textbf{Vehicle 1} & \cellcolor{gray!25}Y: 0.0261 & Y: 0.0102 & Y: 0.0099\\
  (Small) & \cellcolor{gray!25}$\theta$: 0.0331  & $\theta$: 0.0109  & $\theta$: 0.0159\\
  \hline
  \textbf{Model} & X: 0.0354 & \cellcolor{gray!25}X: 0.0145 & X: 0.0303\\
  \textbf{Vehicle 2} & Y: 0.0394 & \cellcolor{gray!25}Y: 0.0176 & Y: 0.0262\\
  (Long) & $\theta$: 0.2007  & \cellcolor{gray!25}$\theta$: 0.0127 & $\theta$: 0.0917 \\
  \hline
  \textbf{Model} & X: 0.0153 & X: 0.0103 & \cellcolor{gray!25}X: 0.0190  \\
  \textbf{Vehicle 3} & Y: 0.0162 & Y: 0.0095 & \cellcolor{gray!25}Y: 0.0182 \\
  (Large) & $\theta$: 0.0666  & $\theta$: 0.0097  & \cellcolor{gray!25}$\theta$: 0.0241\\
  \hline
  \textbf{Model} & \cellcolor{gray!75}X: 0.0080 & \cellcolor{gray!75}X: 0.0087 & \cellcolor{gray!75}X: 0.0083\\
  \textbf{MERGED} & \cellcolor{gray!75}Y: 0.0086 & \cellcolor{gray!75}Y: 0.0087 & \cellcolor{gray!75}Y: 0.0083\\
  (All 3) & \cellcolor{gray!75}$\theta$: 0.0143  & \cellcolor{gray!75}$\theta$: 0.0080 & \cellcolor{gray!75}$\theta$: 0.0115 \\
  \hline
\end{tabular}
\end{table}

\begin{table}[htp]
\centering
\caption{MAEs for the Augmented Buckingham $\pi$ theorem based model (X and Y are in meters and $\theta$ in rad)} 
\label{results_A}
\begin{tabular}{|c|c|c|c|}
  \hline 
  \textbf{Models} & \textbf{Data Vehicle 1} & \textbf{Data Vehicle 2} &\textbf{Data Vehicle 3}\\ 
  & (Small) & (Long) & (Large) \\\hline \hline

  \textbf{Model} & \cellcolor{gray!25}X: 0.0126  & X: 0.0117 & X: 0.0104 \\
  \textbf{Vehicle 1} & \cellcolor{gray!25}Y: 0.0121 & Y: 0.0090 & Y: 0.0077\\
  (Small) & \cellcolor{gray!25}$\theta$: 0.0131  & $\theta$: 0.0043  & $\theta$: 0.0063\\
  \hline
  \textbf{Model} & X: 0.0271 & \cellcolor{gray!25}X: 0.0103 & X: 0.0219\\
  \textbf{Vehicle 2} & Y: 0.0286 & \cellcolor{gray!25}Y: 0.0134 & Y: 0.0222\\
  (Long) & $\theta$: 0.1407  & \cellcolor{gray!25}$\theta$: 0.0052 & $\theta$: 0.0591 \\
  \hline
  \textbf{Model} & X: 0.0130 & X: 0.0090 & \cellcolor{gray!25}X: 0.0117  \\
  \textbf{Vehicle 3} & Y: 0.0133 & Y: 0.0091 & \cellcolor{gray!25}Y: 0.0129 \\
  (Large) & $\theta$: 0.0389  & $\theta$: 0.0038  & \cellcolor{gray!25}$\theta$: 0.0098\\
  \hline
  \textbf{Model} & \cellcolor{gray!75}X: 0.0045 & \cellcolor{gray!75}X: 0.0055 & \cellcolor{gray!75}X: 0.0046\\
  \textbf{MERGED} & \cellcolor{gray!75}Y: 0.0050 & \cellcolor{gray!75}Y: 0.0061 & \cellcolor{gray!75}Y: 0.0051\\
  (All 3) & \cellcolor{gray!75}$\theta$: 0.0060  & \cellcolor{gray!75}$\theta$: 0.0028 & \cellcolor{gray!75}$\theta$: 0.0035 \\
  \hline
\end{tabular}
\end{table}

Table \ref{comparison} summarizes the three previous tables (\ref{results_P}, \ref{results_B} and \ref{results_A}). For self-predictions, we can see that the Buckingham $\pi$ theorem based model and the Augmented Buckingham model are respectively 1.93 times and 3.60 times more precise than the traditional model. These numbers climb to 11.76 times and 15.80 times for cross-predictions. Finaly, the Buckingham $\pi$ theorem based model and the Augmented Buckingham model are respectively 4.80 times and 9.17 times more precise than the traditional model in shared-predictions.

\begin{table}[htp]
\centering
\caption{MAEs comparison for all models trained with simulated data (X and Y are in meters and $\theta$ in rad)} 
\label{comparison}
\begin{tabular}{|c|c|c|c|}
  \hline 
  \textbf{Prediction} & \textbf{Traditional} & \textbf{Buckingham} &\textbf{Augmented}\\ 
  \textbf{types} & \textbf{dimensionalized} & \textbf{$\pi$ theorem} & \textbf{Buckingham} \\
  & \textbf{model} & \textbf{based model} & \textbf{$\pi$ model}\\ \hline \hline

  \textbf{SELF} & \cellcolor{gray!25}X: 0.0408  & \cellcolor{gray!25}X: 0.0180 & \cellcolor{gray!25}X: 0.0115 \\
  \textbf{Predictions} & \cellcolor{gray!25}Y: 0.0362 & \cellcolor{gray!25}Y: 0.0206 & \cellcolor{gray!25}Y: 0.0128\\
   & \cellcolor{gray!25}$\theta$: 0.0416  & \cellcolor{gray!25}$\theta$: 0.0233  & \cellcolor{gray!25}$\theta$: 0.0094\\
  \hline
  \textbf{CROSS} & X: 0.2654 & X: 0.0197 & X: 0.0155\\
  \textbf{Predictions} & Y: 0.2198 & Y: 0.0186 & Y: 0.0128\\
   & $\theta$: 0.6594  & $\theta$: 0.0659 & $\theta$: 0.0422 \\
  \hline
  \textbf{SHARED} & \cellcolor{gray!75}X: 0.0471 & \cellcolor{gray!75}X: 0.0083 & \cellcolor{gray!75}X: 0.0049  \\
  \textbf{Predictions} & \cellcolor{gray!75}Y: 0.0444 & \cellcolor{gray!75}Y: 0.0085 & \cellcolor{gray!75}Y: 0.0054 \\
   & \cellcolor{gray!75}$\theta$: 0.0397  & \cellcolor{gray!75}$\theta$: 0.0113  & \cellcolor{gray!75}$\theta$: 0.0041\\
  \hline

\end{tabular}
\end{table}

\subsubsection{Learning rate} 
\begin{figure*}[htb]
\begin{center}
\includegraphics[width=0.99\linewidth]{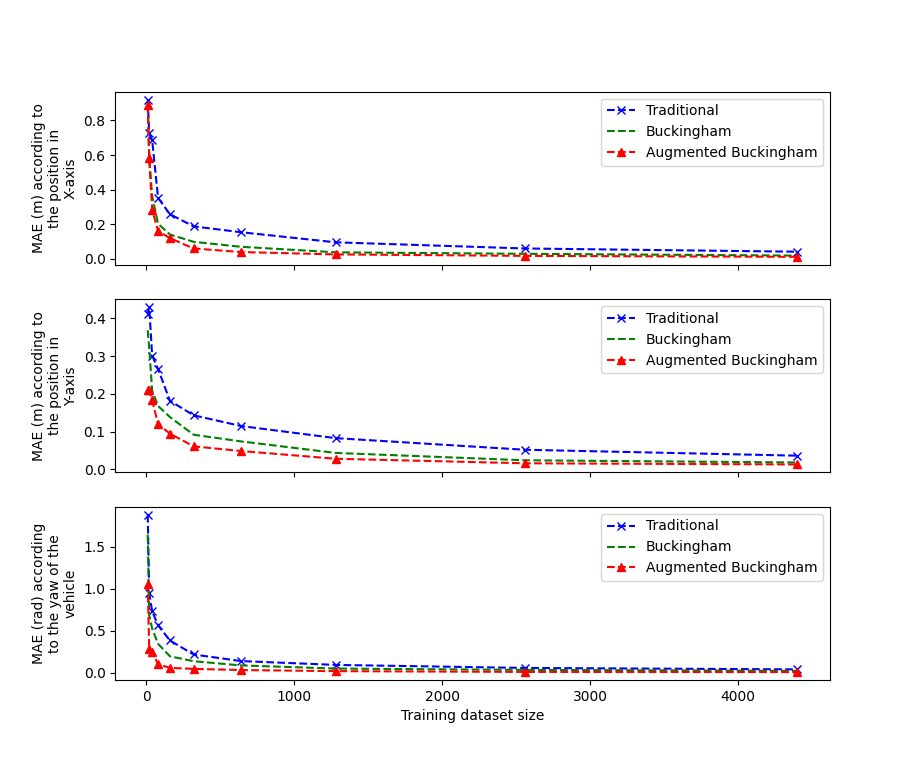}
\caption{MAE convergence according to training size for the simulated data of the large vehicle}\label{Conv}
\end{center}
\end{figure*}

The amount of data required to obtain an operational predictive function for self-predictions for all model types is also analyzed. Figure \ref{Conv} shows the MAE convergence for the final longitudinal position $X$, final lateral position $Y$ and final yaw $\theta$ of the three model types with the large vehicle's self-predictions. For all outputs, the Augmented Buckingham model converges the fastest, followed by the Buckingham $\pi$ theorem based model. More data would be required to achieve full convergence of the final $X$ and $Y$ positions for the traditional dimensionalized model. Only the data for the large vehicle is presented for shortness, as the learning curve for all vehicles were similar.


\subsubsection{Comparative analysis}

\begin{figure*}[htb]
\begin{center}
\includegraphics[width=0.7\linewidth]{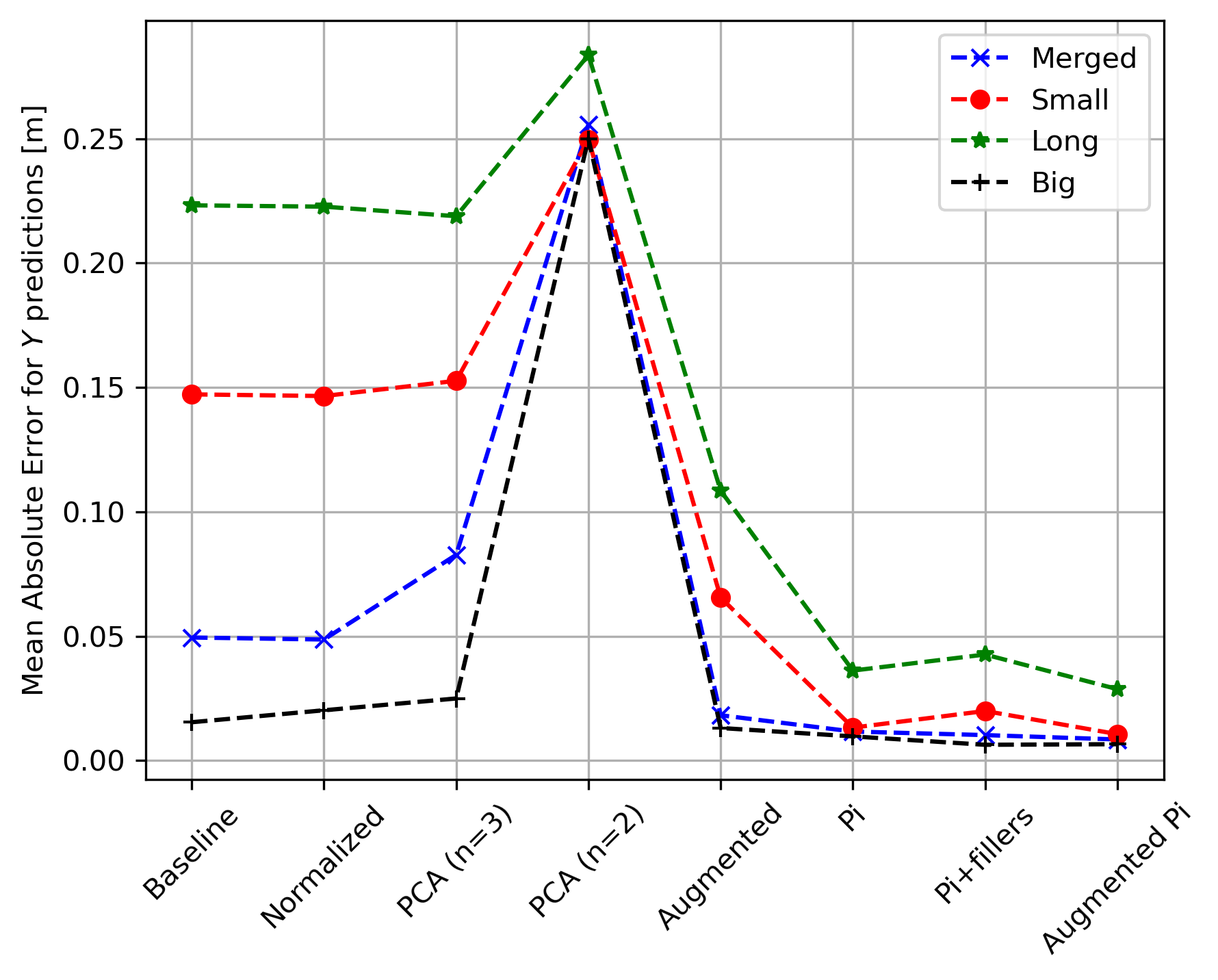}
\caption{Comparison of the performance of various pre-processing scheme for predicting the $Y$ position of the large vehicle. The x-axis indicate the pre-processing scheme used on the data before training the XGBoost algorithm. The four curves represent the source of data used for the training, as indicated in the legend. The green and red curve correspond to transfer learning tasks, i.e. predicting the motion of the large vehicle using only data from other vehicles.}\label{alex_figure}
\end{center}
\end{figure*}

Various hypothesis could explain the performance improvement. \textbf{1) Normalization:} It is possible that the improvement is mainly due to a good normalization on the range of input-output variables which helps the learning algorithm. \textbf{2) Dimensionnality reduction:} It is possible that the improvement is mainly due to the reduction in the numbers of inputs. \textbf{2) Better coordinate system:} It is possible that the improvement is mainly due to the coordinate transformation in the dimensionless space that make the learning easier and the knowledge more generic. In order to better understand the results and the source of improvements, a comparative analysis is conducted with multiple variations to help isolate the main source of improvements. The results are summarized at Figure \ref{alex_figure}.

For this analysis, the learning problem is simplified to predicting only one output: the final $Y$ position of the large vehicle. The performance of each learning scheme will be compared in terms of MAEs of the prediction for a test dataset. In addition to the three scheme already presented, i.e. the baseline, the Buckingham $\pi$ and the augmented Buckingham $\pi$, five new variations are compared. \textbf{Normalized:} The data is first normalized by the largest value seen in the training data (for each input variable). \textbf{PCA:} A principal component analysis (PCA) transformation is conducted on the input data before training the model and making prediction. The results include when 3 or 2 components are used, with 2 corresponding to the number of inputs when using the Buckingham $\pi$ scheme. \textbf{Augmented:} The input data is augmented with an handcrafted feature, corresponding to $\dot{\theta}$ at Equation \eqref{eq:pi6}, i.e. the same as the \textit{Augmented Buckingham $\pi$ scheme} but without the additional step to make it a dimensionless number. \textbf{Pi + fillers:} The two $\pi$ groups of the \textit{Buckingham $\pi$ scheme} are augmented with two original dimensional input variables: $v_i$ and $l$, i.e. superfluous information according to dimensional analysis. 

Results illustrated at Figure \ref{alex_figure}, show that compared to the baseline approach, normalizing the data is having almost no effect on the performance, PCA reduction of the input data is always leading to worst performance, and augmenting the data with $\dot{\theta}$ is always leading to a good performance improvement, but less then when using the Buckingham $\pi$ based schemes. Finally, comparing the three Buckingham $\pi$ based schemes, augmenting the data can increase or decrease the performance, according to the relevance of the additional features.

\subsubsection{Discussion}

As it was expected, the dimensionless models performed much better for generalizing the predictions, i.e. for predicting the motion of a vehicle using data from a single different vehicle (cross-predictions), or a shared database of the 3 vehicles (shared predictions). It is also interesting, even surprising, that there was an improvement in the accuracy of the self-prediction, for which the benefit of using dimensionless variables are less direct. From the comparative study, it seems that there is something fundamentally better with using dimensionless variable as the coordinate system for encoding the input information, since alternate preprocessing schemes did not led to similar improvements. 

Regarding the Augmented Buckingham model, there was a significant improvement compared to the Buckingham model without the hand-crafted input variable. The improvement can be seen mainly early in the learning phase (see Figure \ref{Conv}), but then as more data in coming the performance of the other models is slowly catching up. An hypothesis is that this model as a head-start, already knowing some of the non-linearity (the $\tan$ term) while the other have to learn it from the data. All in all, this experiment is interesting because all the variables involved in generating the data were exactly known. This illustrated how drastic the improvement could be in a perfect situation by leveraging dimensional analysis in the learning scheme.

\section{Learning with experimental data}
\label{sec:exp}

The validation presented in the last section, consisted of a very clean ideal situation. Here, another validation is conducted for a much more challenging learning task. Here the task is to predict much more dynamic braking maneuvers of real small platforms, based on real data recorded with a vision-based acquisition system to provide the position of the vehicles. Hence the problem is harder because, 1) the data is noisy, 2) the behaviour to model is more complex and 3) all relevant variables are not known, some simplifying hypothesis must be made. 

\subsection{Methodology for generating the data}

For all three vehicles, 540 experimental tests were carried out with the initial values and maneuvers shown in table \ref{valuesexp}: three type of terrains, 6 initial velocity levels, 10 levels of braking and 3 steering angles. All tests are conducted indoor on a flat and uniform surface, with various material to modify the friction coefficient. All tests start with the vehicle in position (0,0), orientation 0 rad, initial speed $v_i$ and ground/tire friction coefficient $\mu$. The maneuver begins immediately with a deceleration of the rear wheel given by $a$ (the servo-motor is used to impose an angular deceleration of $a/r$ where $r$ is the wheel radius) and setting the steering angle to position $\delta$. The test stops when the vehicle has come to a complete stop, and the final position and orientation are recorded using a VICON triangulation camera system \cite{vicon}.

\begin{table}[ht]
   \centering 
   \caption{Input values for the experimental tests for all 3 vehicles} 
   \label{valuesexp}
   \begin{tabular}{c c c c}
   \hline 
   \hline \noalign{\smallskip} \noalign{\smallskip} \noalign{\smallskip}
    $\mu$ & $v_i$ & $a$ & $\delta$\\&(m/s)&g (9.81 m/s$^2$)& (rad)\\ \noalign{\smallskip} \noalign{\smallskip} \hline \hline \noalign{\smallskip} 
   0.2 & From 1.0 & From 0.1 & 0.0000\\ 
   0.4 & to 3.5 & to 1.0 & 0.3927 \\
   0.9 & by 0.5 & by 0.1 & 0.7854 \\\noalign{\smallskip} 
   \hline\noalign{\smallskip} 
   3 values & 6 values & 10 values & 3 values \\\noalign{\smallskip} 
   \noalign{\smallskip} \hline \hline \noalign{\smallskip}
   \end{tabular}
\end{table}

\subsection{Model for the dynamic maneuvers}

To capture some of the dynamic behavior of the dynamic maneuvers, more input variables are considered in the learned predictor function here. The relevant variable to include, were sectioned with the assumption that a dynamic bicycle model (often used in planing and control pipeline of autonomous robots \cite{paden_survey_2016}) would be a good approximation of the behaviour. Additionally, the tire-ground interaction is assumed to be capture by constant friction coefficient $\mu$ throught the maneuver. Hence, all the states, inputs and vehicle parameters that appear in those equations are used to form the input of the predictor model in this section, see Table \ref{expVari}.

\subsubsection{Traditional baseline model}

\begin{figure}[ht]
\vspace{-10pt}
\begin{center}
\includegraphics[width=0.99\linewidth]{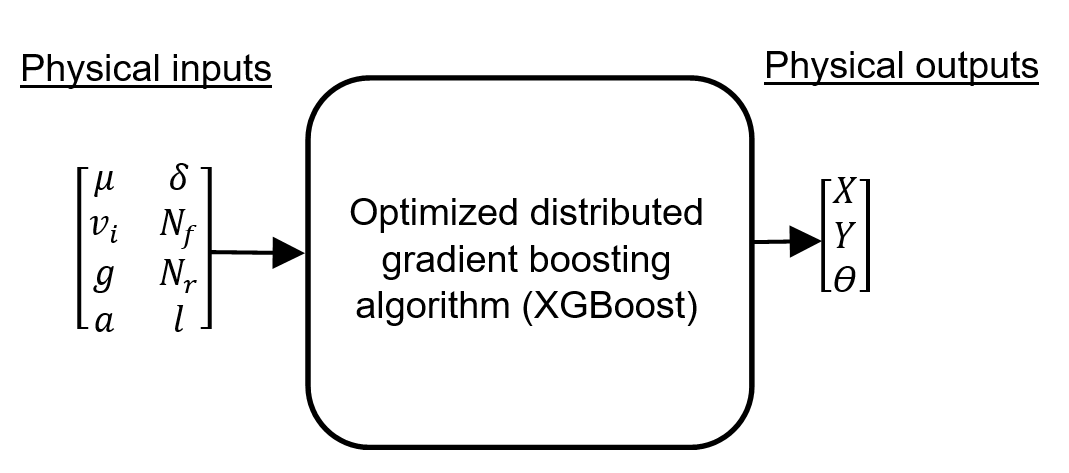}
\caption{Baseline learning scheme for experimental data}\label{phy_filter}
\end{center}
\vspace{-10pt}
\end{figure}

The process of integrating the nonlinear equations describing the dynamic bicycle model until the vehicle comes to a stop, can be written in the following input-relationship form:
\begin{equation}\label{eq:dyn}
\begin{bmatrix}
X \\
Y \\
\theta
\end{bmatrix}
=
f \Bigl( \mu, v_i, g, a, \delta, N_f, N_r , l \Bigr)
\end{equation}
The learned model for the dynamic maneuver will thus be based on using the same input variables to predict the final position obtain for the real experimental maneuver. Here it is important to note that this will be an approximation: some variables that might have some influence on the outcome (for instance the suspension rigidity) are not included as available inputs for the model. The baseline learning scheme for the experimental validation is, as illustrated in Figure \ref{phy_filter}, using directly the XGBoost algorithm to learn an data-driven approximation to equation \eqref{eq:dyn} relationship.

\subsubsection{Buckingham $\pi$ theorem based model}

\begin{figure*}[ht]
\begin{center}
\includegraphics[width=0.7\linewidth]{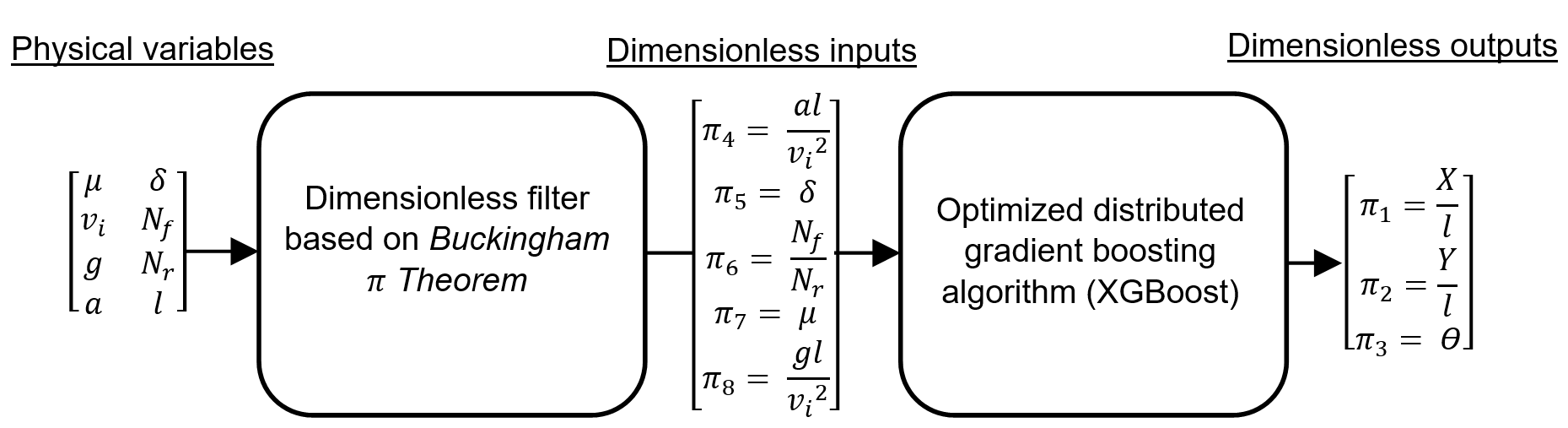}
\caption{Buckingham $\pi$ theorem based model for experimental data}\label{adim_filter}
\end{center}
\vspace{-10pt}
\end{figure*}

Here if we conduct the dimensional analysis, each line of the relationship of equation \eqref{eq:dyn} involves $N$ = 9 variables and \textit{P} = 3 independent dimensions (mass [M], length [L] and times [T]). Then, based on the Buckingham $\pi$ theorem, each relationship (lines of equation \eqref{eq:dyn}) could be restated as the relationship between 6 dimensionless groups, as follow:
\begin{equation}\label{eq:dyn2}
\begin{bmatrix}
\pi_1 \\ \pi_2 \\ \pi_3
\end{bmatrix}
=
f(\pi_4, \pi_5, \pi_6, \pi_7, \pi_8)
\end{equation}
By using the distance between the two axles $l$ [M], the initial velocity of the vehicle $v_i$ [MT$^{-1}$] and the normal force on front wheels $N_f$ [MLT$^{-2}$] as repeated variables, the following eight dimensionless numbers were used:
\begin{align}
\pi_1 = \frac{X}{l} \quad 
\pi_2 = \frac{Y}{l} \quad 
\pi_3 = \theta \quad 
\pi_4 = \frac{al}{v_i^2} \\
\pi_5 = \delta  \quad 
\pi_6 = \frac{N_f}{N_r} \quad 
\pi_7 = \mu \quad 
\pi_8 = \frac{gl}{v_i^2}
\end{align}
The new dimensionless relationship to predict is thus in the following form:
\begin{align}\label{fonc_sim_dim}
\begin{bmatrix}
\frac{X}{l} \\ \frac{Y}{l} \\ \theta
\end{bmatrix}
&=
f \left(\frac{al}{v_i^2}, \delta , \frac{N_f}{N_r} , \mu , \frac{gl}{v_i^2}\right)
\end{align}
and the overall pipeline for making prediction with the dimensionless approach is shown in figure \ref{adim_filter}.

\subsubsection{Augmented Buckingham $\pi$ theorem based model for experimental data}

Here, to test if handcrafted feature would be advantageous, two additional dimensionless numbers are chosen for their dynamic significance.
Two ratios representing adherence limits for the longitudinal force $\pi_9$ and the lateral force $\pi_{10}$ are introduced as described by equations \ref{long_cor} and \ref{lat_cor}:

\begin{equation}\label{long_cor}
\pi_{9} = \frac{F_{limit}}{F_{x_{maneuver}}} = \frac{N_r\mu}{ma} = \frac{N_r\mu g}{(N_f+N_r)a}
\end{equation}
\begin{equation}\label{lat_cor}
\pi_{10} = \frac{F_{limit}}{F_{y_{maneuver}}} = \frac{N_r\mu}{\frac{mv^2}{R}} = \frac{g\mu R}{v^2} = \frac{g\mu l}{v^2\tan\delta}
\end{equation}

\begin{figure*}[ht]
\begin{center}
\includegraphics[width=0.8\linewidth]{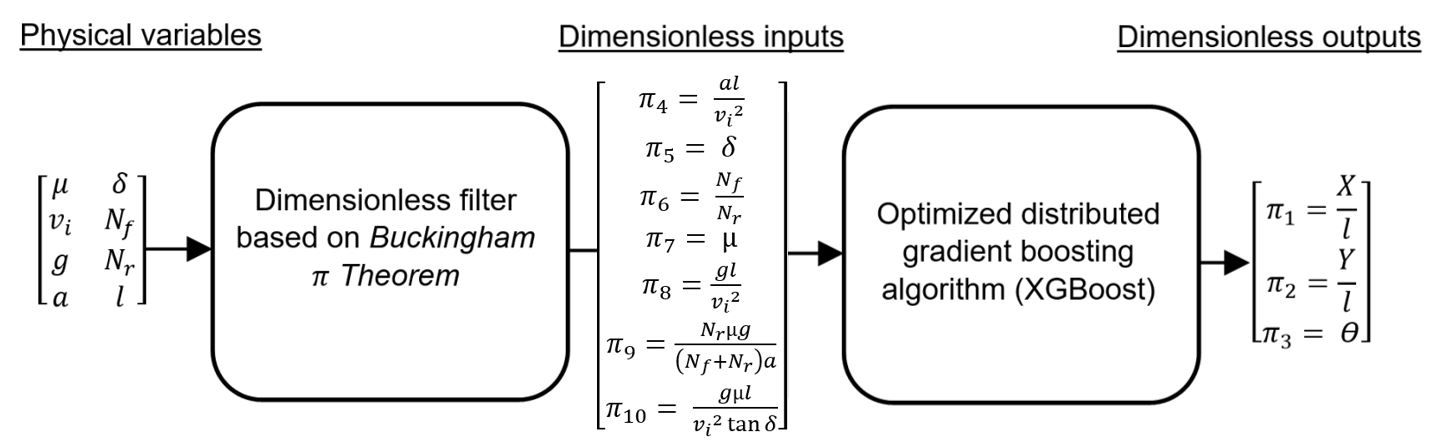}
\caption{Augmented Buckingham $\pi$ theorem based model for experimental data}\label{AM_exp}
\end{center}
\vspace{-10pt}
\end{figure*}

The new dimensionless relationship to predict is thus in the following form:
\begin{align}\label{fonc_sim_dim}
\begin{bmatrix}
\frac{X}{l} \\ \frac{Y}{l} \\ \theta
\end{bmatrix}
&=
f \left(\frac{al}{v_i^2}, \delta , \frac{N_f}{N_r} , \mu , \frac{gl}{v_i^2}, \frac{N_r\mu g}{(N_f+N_r)a}, \frac{g\mu l}{v^2\tan\delta} \right)
\end{align}
and the overall pipeline for making prediction with the dimensionless approach is shown in figure \ref{AM_exp}.

\subsection{Results}

As for the simulation data, the results for each model type are examined from two perspectives: 1) the accuracy (MAEs) and 2)the learning rates for each situations.

\subsubsection{Mean Absolute Error} To avoid any redudancy, only the summarizing table \ref{comparison_exp} for self, cross and shared-predictions is presented for the experimental data. For self-predictions, we can see that the Buckingham $\pi$ theorem based model and the Augmented Buckingham model are respectively 1.11 times and 1.14 times more precise than the traditional model on average. These numbers drop to 1.05 times and 1.07 times more precise for cross-predictions. Finaly, the Buckingham $\pi$ theorem based model and the Augmented Buckingham model are respectively 1.11 times and 1.28 times more precise than the traditional model in shared-predictions.

\begin{table}[ht]
\centering
\caption{MAE comparison for all models trained with experimental data (X and Y are in meters and $\theta$ in rad)} 
\label{comparison_exp}
\begin{tabular}{|c|c|c|c|}
  \hline 
  \textbf{Prediction} & \textbf{Traditional} & \textbf{Buckingham} &\textbf{Augmented}\\ 
  \textbf{types} & \textbf{dimensionalized} & \textbf{$\pi$ theorem} & \textbf{Buckingham} \\
  & \textbf{model} & \textbf{based model} & \textbf{$\pi$ model}\\ \hline \hline

  \textbf{SELF} & \cellcolor{gray!25}X: 0.1229  & \cellcolor{gray!25}X: 0.1228 & \cellcolor{gray!25}X: 0.1197 \\
  \textbf{Predictions} & \cellcolor{gray!25}Y: 0.0488 & \cellcolor{gray!25}Y: 0.0321 & \cellcolor{gray!25}Y: 0.0335\\
   & \cellcolor{gray!25}$\theta$: 0.0482  & \cellcolor{gray!25}$\theta$: 0.0596  & \cellcolor{gray!25}$\theta$: 0.0554\\
  \hline
  \textbf{CROSS} & X: 0.3027 & X: 0.3752 & X: 0.3775\\
  \textbf{Predictions} & Y: 0.2198 & Y: 0.0186 & Y: 0.0128\\
   & $\theta$: 0.2007  & $\theta$: 0.1494 & $\theta$: 0.1454 \\
  \hline
  \textbf{SHARED} & \cellcolor{gray!75}X: 0.0539 & \cellcolor{gray!75}X: 0.0507 & \cellcolor{gray!75}X: 0.0400  \\
  \textbf{Predictions} & \cellcolor{gray!75}Y: 0.0130 & \cellcolor{gray!75}Y: 0.0105 & \cellcolor{gray!75}Y: 0.0089 \\
   & \cellcolor{gray!75}$\theta$: 0.0193  & \cellcolor{gray!75}$\theta$: 0.0187  & \cellcolor{gray!75}$\theta$: 0.0187\\
  \hline

\end{tabular}
\end{table}

\subsubsection{Learning rate} 
\begin{figure*}[htbp]
\begin{center}
\includegraphics[width=0.99\linewidth]{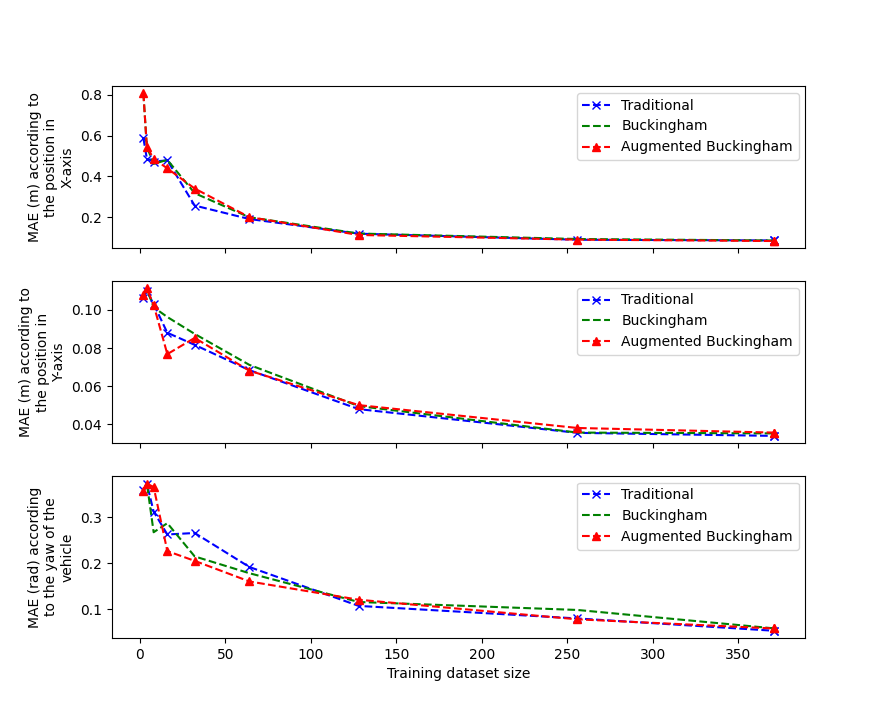}
\caption{MAE convergence according to training size for the experimental data of the large vehicle}\label{fig:Conv_EXP}
\end{center}
\end{figure*}

 Figure \ref{fig:Conv_EXP} shows the MAE convergence for the final longitudinal position $X$, final lateral position $Y$ and final yaw $\theta$ of the three model types with the large vehicle's experimental self-predictions. For all outputs, all three models follow a similar learning curve. More training data (limited here by the number of experimental test conducted) would be required to fully converge since all curves keep on improving at a relatively high rate.

\subsubsection{Discussion}

Here it is interesting to note that even for the much harder realistic learning task, the proposed Buckingham $\pi$ schemes still led to always improving the prediction accuracy when using a shared database from the data of the 3 vehicles. However, the improvements were found to be less spectacular then for the clean simulated data. This was expected since for this experiment: 1) the data was noisy (the main source would be the vision-based measurement system), 2) the input-output relationship to predict is more complex (mainly the complex tire-ground relationship), and 3) the given list of relevant inputs to the model is not a complete list of variables that influence the outcome (unlike previously when the data was generated by a known controlled simulation model), it was only an approximation based on the dynamic bicycle model equations. In future works it would be interesting to conduct more intermediary experiments to isolate the effect of each difficulties individually.


\section{Conclusion}

 To conclude, this paper presents a dimensionless learning scheme designed to allow the sharing of data and models between cars of various kinematic and dynamic parameters, and results from a case-study demonstrating the performance of the scheme to generalize in comparison to a baseline learning approach. The results of the case study based on kinematic simulations show that the proposed dimensionless models are much more accurate when using a shared database (an order of magnitude more accurate), and have faster learning rates than a traditional dimensional model. The results of the case study based on experimental data of car-like robots conducting dynamic braking maneuvers also showed an advantage with the proposed dimensionless scheme over a baseline approach (in the range of 5\% to 30\% of precision improvements). The gain in the experimental context was less significant than what was observed through simulations, which is mainly attributable to a discrepancy between the assumptions made (i.e. when selecting the list of input variables for the dimensional analysis) and the high complexity of the tire dynamics involved in the experiments. However, it is interesting to note that using a dimensionless relationship for the predictions always led to improvements in all the tested scenario. Since it is an easy prepossessing step to conduct it should be used all the time, especially since if all relevant variables are included this a lossless technique to reduce the dimension of the input and output space according to the  Buckingham $\pi$ theorem. In future work, it would be interesting to conduct more tests in intermediary situations to better map the potential improvements one could expect by using the presented dimensionless approach. 

All in all, results presented in this case-study have shown that leveraging Buckingham $\pi$ theorem can lead to very interesting improvements in terms of experimental data required to train a learning algorithm modelling a vehicle, to transfer results between similar vehicles and when using a shared database. Since the requirements for using dimensional analysis and applying the Buckingham $\pi$ theorem are simply that the relationship (to predict) is involving physically meaningful variables and knowing which one, we should expect similar results for generalizing a learned model involving various type of vehicles, robotics systems, or in fact any physically meaningful non-linear relationship to predict. However, there are still many unanswered questions regarding this type of learning technique. Hence, the approach should be investigated further to better understand its full potential and limitations.

\bibliographystyle{IEEEtran}
\bibliography{zotero_old}

\end{document}